\documentclass[runningheads]{llncs}
\usepackage{graphicx}
\usepackage{booktabs}
\usepackage{array}
\usepackage{amsmath}
\usepackage{color,soul}
\usepackage[table]{xcolor}
\usepackage[linesnumbered,ruled,vlined]{algorithm2e}
\usepackage{lscape}
\usepackage[skins]{tcolorbox}
\usepackage{hyperref}

\usepackage{bbding}
\usepackage{adjustbox}

\begin{document}
\title{Stratified Sampling for Extreme Multi-Label Data}
%
%
%
\author{Maximillian Merrillees and Lan Du\Envelope}
\authorrunning{Maximillian Merrillees and Lan Du}
%
\institute{Faculty of Information Technology\\ Monash University, Clayton VIC 3800, Australia\\
\email{maxi.merrillees@gmail.com},\,
\email{lan.du@monash.edu}
}

%
%
\maketitle              

\begin{abstract}
Extreme multi-label classification (XML) is becoming increasingly relevant in the era of big data. 
Yet, there is no method for effectively generating stratified partitions of XML datasets. 
Instead, researchers typically rely on provided test-train splits that, 
1) aren't always representative of the entire dataset, and 2) are missing many of the labels. 
This can lead to poor generalization ability and unreliable performance estimates, 
as has been established in the binary and multi-class settings. 
As such, this paper presents a new and simple algorithm that can efficiently generate 
stratified partitions of XML datasets with millions of unique labels. 
We also examine the label distributions of prevailing benchmark splits, 
and investigate the issues that arise from using unrepresentative subsets of data for model development. 
The results highlight the difficulty of stratifying XML data, 
and demonstrate the importance of using stratified partitions for training and evaluation.

\end{abstract}
\section{Introduction}
The composition of data used for training and testing can have a big impact on the model development process. It can influence choices regarding training strategy and hyperparameter selection, and can also effect performance estimates. As such, for classification tasks, stratified sampling is commonly used to generate these subsets because they have been shown to result in performance estimates with lower bias and variance compared to random sampling~\cite{ref_article_bias}.

Performing stratified sampling on binary and multi-class datasets is straightforward. Each data point is only associated with one label, so generating stratified splits can be achieved by simply sampling instances of each class based on its prevalence in the dataset. On the other hand, generating stratified subsets of multi-label data is more difficult because each instance can be associated with one or more labels. Assigning an instance to a subset based on one of its labels will impact all the other labels associated with that instance.

As such, to facilitate comparability between models, it's common for XML researchers to use benchmark datasets with provided test-train splits~\cite{ref_article_parabel}~\cite{ref_article_fastxml}~\cite{ref_article_annexml}~\cite{ref_article_attentionxml}. However, our investigation into these provided splits revealed that a number of them have sub-optimal characteristics. Specifically, there are subsets where the distribution of labels is very dissimilar to that of the entire dataset. Also, there are test sets where a significant proportion of rare labels are missing. 

This can be problematic since using unrepresentative data can lead to biased results or sub-optimal choices for a model's hyperparameters. 
Also, excluding a large proportion of rare labels from being evaluated implies that a model's performance on rare labels is not important. That's not always the case. 
In fact, the authors of PfastreXML argue that such situations are common~\cite{ref_article_pfastrexml}. They provide the example of tagging Wikipedia articles, where there would be less value in assigning a generic label like {\textit{Poem}} than novel labels like {\textit{Epic poems in Italian}} or {\textit{14th century Christian texts}}. For another example, in the medical domain, some diseases are rare. Therefore, the corresponding ICD labels might not be observed as often as some other common diseases, like seasonal influenza. However, less frequent diseases do not mean they are not important.

These issues have been known for some time in multi-label classification (MLC) research. A paper from 2011 points out similar issues, and presented an iterative algorithm to perform stratified partitioning of MLC datasets~\cite{ref_article_iterative}. Since then, one other method has been proposed~\cite{ref_article_network}. A review of recent literature suggests that the iterative algorithm from 2011 remains the preferred stratification method~\cite{ref_article_lince}~\cite{ref_article_mlc1}~\cite{ref_article_mlc2}. However, our tests of the iterative algorithm revealed that it was unable to effectively generate stratified splits of XML data. This is possibly due to the difference in scale between MLC and XML data. In any case, there is currently no method, to our knowledge, for efficiently generating stratified test-train splits of XML datasets. Given the increasing prevalence of XML research, this gap needs to be addressed.

In this paper, we propose a new stratified sampling algorithm for XML data, which makes use of a simple sampling strategy that swaps instances between the train and the test sets based on scores that measure the dissimilarity between the current split state and the stratified state. Compared with random sampling and iterative sampling, the proposed algorithm: 1) achieves lower KL divergence between the label distribution of each test set against the distribution of their respective data set, and 2) is much faster than the iterative sampling method that is widely used in multi-label learning. Meanwhile, we also examine the label distributions of prevailing benchmark splits, and highlight the bias in performance estimates that can arise from using unrepresentative subsets of data. 

This paper is structured as follows: Section 2 provides a summary of related works. Section 3 contains an overview of XML datasets. We present our algorithm in Section 4, and the resulting partitions in Section 5. In Section 6, we highlight the difference in performance estimates from using different splits. Finally, we provide our concluding remarks in Section 7.

\section{Related Works}

The first method for generating stratified partitions of MLC data was presented by Sechidis and Tsoumakas in 2011~\cite{ref_article_iterative}. Their iterative algorithm creates k-subsets of data by allocating data points to subsets one-by-one based on the suitability of its labels. Prior to that algorithm, MLC researchers typically performed random sampling or relied on the splits available through the MULAN repository~\cite{ref_article_mulan}.

The authors were motivated to develop such an algorithm because it had been established that using stratified partitions results in performance estimates with lower bias and variance compared to estimates obtained using randomly sampled data~\cite{ref_article_bias}. Yet, up to that point, no stratification method existed for MLC data. In their paper, they demonstrated that using stratified subsets resulted in more robust performance estimates. A python implementation of the algorithm is available from the scikit-multilearn package~\cite{ref_article_multilearn}.

Since then, one other stratification method has been proposed for MLC datasets. In 2017, an extension of the iterative algorithm was presented that takes into account second-order relationships between labels~\cite{ref_article_network}. That is, an algorithm that seeks to maintain the distribution of label-pairs during partitioning. In their paper, the authors demonstrated how doing so improves classification performance as measured by label-pair oriented metrics. Despite this newer method, a review of the recent literature suggests that the iterative algorithm from 2011 remains the preferred stratified partitioning method. Perhaps this is because second-order relationships are not typically considered in MLC research.

Despite the widespread use of the iterative algorithm, this study found it unsuitable for large-scale XML datasets. The algorithm is slow, and does not seem to be capable of generating well stratified partitions for XML datasets. This is not surprising given the method was developed and tested on relatively small MLC datasets. Indeed, the largest dataset they considered had just 983 labels and 16K data points - much smaller than XML data that can contain millions of labels and data points.

\section{Overview of XML Datasets}

XML is defined by datasets that contain thousands to millions of labels - this is what makes it \textit{extreme}. Another notable characteristic is the high proportion of labels with few associated instances - typically referred to as tail labels\footnote{For this paper, tail labels are those with fewer than 10 instances in the dataset.}. Also, as with MLC datasets in general, each data point is associated with multiple labels. Table~\ref{tab_datastats} summarizes the basic statistics of four datasets that are often used in XML research, the label size of which ranges from 4K to 670K, which often follows a power-law distribution \cite{lu2020multi}.

Together, these properties make it challenging to create stratified partitions. Having multiple labels per data point means conventional stratification methods cannot be applied, and the high proportion of rare labels makes random sampling risky, since it's possible to generate subsets with missing labels. Finally, the large output space makes iterative partitioning slow. Currently, to our knowledge, there does not exist a stratification method that can overcome these problems.

\begin{table}[!t]
  \centering
  \caption{Statistics for a selection of XML datasets~\cite{ref_url_xmlrepo}}
    \def\arraystretch{1.2}%
    \begin{adjustbox}{width=\textwidth}
    \begin{tabular}{p{8em}
    >{\centering\arraybackslash}p{4.5em}
    >{\centering\arraybackslash}p{4.5em}
    >{\centering\arraybackslash}p{4.5em}
    >{\centering\arraybackslash}p{6.5em}
    >{\centering\arraybackslash}p{6.7em}
    >{\centering\arraybackslash}p{4.5em}}
    \toprule
     & \textbf{Num Labels} & \textbf{Num Train} & \textbf{Num Test} & \textbf{Avg. Labels per Sample} & \textbf{Avg. samples per Label} & \textbf{\% Tail Labels} \\
    \midrule
    \rowcolor{gray!10} EURLex-4K & 3,993 & 15,539 & 3,809 & 5.31 & 25.73 & 59\% \\
    Wiki10-31K & 30,938 & 14,146 & 6,616 & 18.64 & 8.52 & 83\% \\
    \rowcolor{gray!10} Delicious-200K & 205,443 & 196,606 & 100,095 & 75.54 & 72.29 & 51\% \\
    Amazon-670K & 670,091 & 490,449 & 153,025 & 5.45 & 3.99 & 89\% \\
    \bottomrule
    \end{tabular}
    \end{adjustbox}
  \label{tab_datastats}%
\end{table}%

\begin{figure}[!t]
\includegraphics[width=1.05\textwidth]{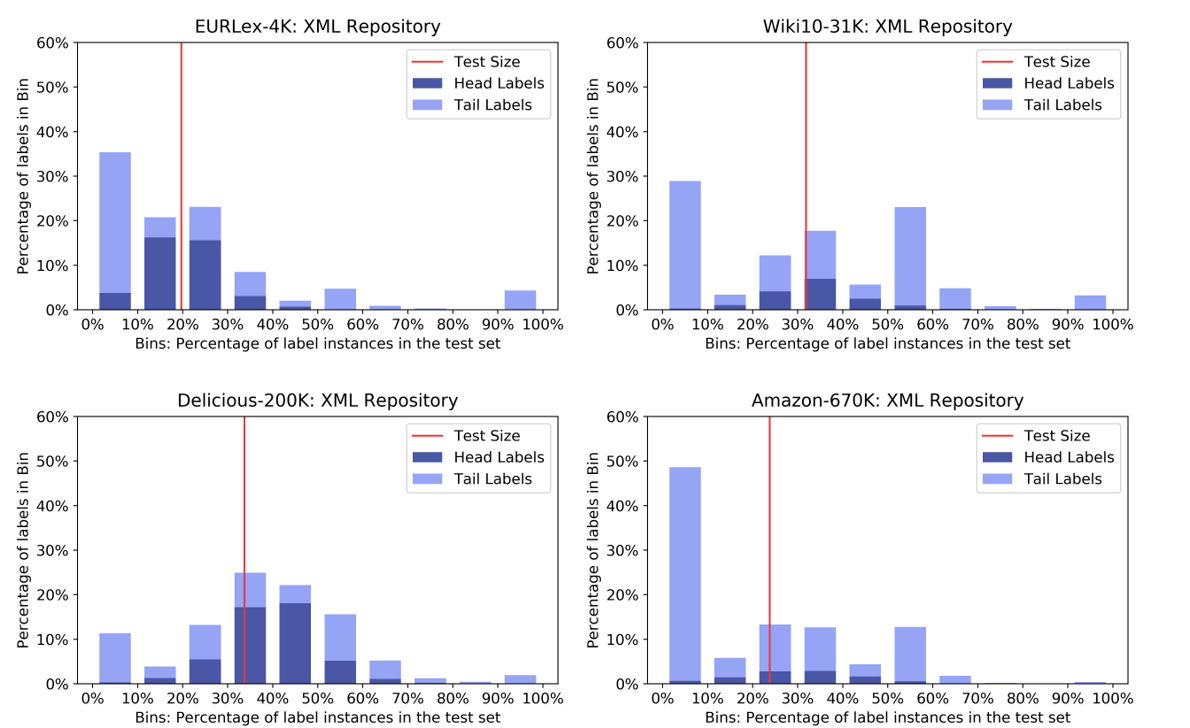}
\caption{Label distributions of provided splits from the XML repository} \label{fig_predetermiendsplits}
\end{figure}

Given the lack of a suitable sampling method, XML researchers typically use test-train splits provided by the XML repository~\cite{ref_url_xmlrepo}. Doing so negates the need to generate their own partitions, and ensures that performance estimates are comparable between papers. However, our examination of these splits reveals that they aren't always representative of their datasets, particularly, there exist many labels that do not have any testing instances.
This is shown in Figure~\ref{fig_predetermiendsplits}, where each chart shows the proportion of labels that fall in each of the 10 bins, where each bin represents a range of the proportion of label instances that appears in the test set. For example, for EURLex-4K, 36\% of the labels have between 0\% and 10\% (first bin) of their instances present in the test set - the remaining labels are spread out across the other 9 bins. The red vertical line represents the test size as a proportion of the entire dataset. In a perfectly stratified test set, all the labels would have this percentage of instances in the test set. Categorizing the labels as either a head- or tail- label reveals that it's mostly tail labels that fall within the first and final bins.

A closer examination reveals that almost all labels that fall within the first bin are actually completely missing from the test set. The proportion missing labels is provided in Table~\ref{tab_missinglabels}. This is problematic because it means that models trained on these splits aren't being adequately evaluated on their performance on rare labels. It is also known that models' performance can be largely impacted by those labels, known as few-shot learning \cite{lu2020multi}. This paper also points out the following conundrum: why train for labels that never get tested? 

\begin{table}[!t]
  \centering
  \caption{Percentage of labels missing from the provided test set}
    \begin{tabular}{
    >{\centering\arraybackslash}p{8.5em}
    >{\centering\arraybackslash}p{8.5em}
    >{\centering\arraybackslash}p{8.5em}
    >{\centering\arraybackslash}p{8.5em}}
    \toprule
    {\textbf{EURLex-4K}} & {\textbf{Wiki10-31K}} & {\textbf{Delicious-200K}} & {\textbf{Amazon-670K}} \\
    \midrule
    32.4\% & 28.7\% & 10.4\% & 48.2\% \\
    \bottomrule
    \end{tabular}%
  \label{tab_missinglabels}%
\end{table}%

\section{Stratified Sampling Algorithm}

In this Section, we present our stratified sampling algorithm. The inputs to the algorithm are the set of documents, $X$, the set of associated labels, $y$, and the $target\_test\_size$.
It generates $X\_train$, $X\_test$, $y\_train$, and $y\_test$, much in the same way as the train\_test\_split() function from scikit-learn~\cite{ref_article_scikitlearn}. Documentation for the python implementation of the stratified sampling algorithm is available on GitHub\footnote{\url{https://github.com/maxitron93/stratified_sampling_for_XML}}. The pseudo-code for the algorithm is outlined in Algorithm 1. Essentially, it seeks to minimize a score that measures how far the current state is from a well-stratified state. A high positive score indicates the partitions are far from stratified. Scores close to 0 indicate that the partitions are well stratified.

The algorithm starts by randomly allocating each data point to the train or test set so the $target\_test\_size$ is achieved (line 1). Then, it performs stratified sampling for a number of epochs. First, it counts the instances of each label in each partition (lines 6 to 10) and calculates a score for each label based on the extent to which a label's $actual\_test\_proportion$ diverges from the $target\_test\_size$ (lines 12 to 17). Scores are normalized to be within the range of 0 and $\pm$1. A positive score means too much of the label is in the test set, while a negative score means too much of the label is in the train set. For example, if a the $target\_test\_size$ is 20\%, and 60\% of a label's instances are in the test set, the label score would be 0.5, since 60\% is half-way between 20\% and 100\%. If a label's $actual\_test\_proportion$ is 5\%, the label score would be -0.75, since 5\% is 75\% of the way between 0\% and 20\%.

After calculating a score for each label, it calculates a score for each data point based on the scores of its labels (lines 19 to 27). A high data point score indicates that many of its labels have too many instances in the data point's current partition, and the datapoint should be swapped to the alternate partition. 

\begin{tcolorbox}[blanker,float=tbp, grow to left by=0.6cm,grow to right by=0.6cm]
\begin{algorithm}[H]
\caption{Stratified sampling algorithm}
\SetAlgoLined
// Start by randomly allocating each data point to either $X\_train$ or $X\_test$ so the test size is equal to $target\_test\_size$ \\
\hfill\\
// Perform stratified sampling for 50 epochs\\
\SetKwBlock{While}{}{}\textbf{while} $epoch$ $\leq$ $50$ \textbf{do}
\While{
    \hfill\\
    // For each $label$, count the appearances in the train and test sets\\
    \SetKwBlock{For}{}{}\textbf{for} $x$ in $X$ \textbf{do}
    \For{
        \SetKwBlock{For}{}{}\textbf{for} $label$ in $label\_set$ of $x$ \textbf{do}
        \For{
            \lIf{$x$ in $X\_train$}{$label\_train\_count$ += $1$}
            \lElse{$label\_test\_count$ += $1$}
        }
    }
    \hfill\\
    // Calculate the score for each $label$\\
    \SetKwBlock{For}{}{}\textbf{for} $label$ in $y$ \textbf{do}
    \For{
        \lIf{$actual\_test\_proportion$ $\geq$ $target\_test\_size$}{\\~~~~$label\_score$ = $\frac{actual\_test\_proportion - target\_test\_size}{1 - target\_test\_size}$}
        \lElse{\\~~~~$label\_score$ = $\frac{actual\_test\_proportion - target\_test\_size}{target\_test\_size}$}
    }
    \hfill\\
    // Calculate the score for each data point\\
    \SetKwBlock{For}{}{}\textbf{for} $x$ in $X$ \textbf{do}
    \For{
        \SetKwBlock{For}{}{}\textbf{for} $label$ in $label\_set$ of $x$ \textbf{do}
        \For{
            \SetKwBlock{If}{}{}\textbf{if} $label\_score > 0$ \textbf{then}
            \If{
                \lIf{$x$ in $test$}{$data\_point\_score$ += $label\_score$}
                \lElse{$data\_point\_score$ -= $label\_score$}
            }
            \SetKwBlock{Else}{}{}\textbf{else}
            \Else{
                \lIf{$x$ in $train$}{$data\_point\_score$ -= $label\_score$}
                \lElse{$data\_point\_score$ += $label\_score$}
            }
        }
    }
    \hfill\\
    // Calculate threshold score for swapping\\
    $threshold\_score$ = percentile($scores$, $threshold\_proportion$)\\
    \hfill\\
    // Swap proportion of data points with high scores to the alternate partition\\
    \SetKwBlock{For}{}{}\textbf{for} $x$ in $X$ \textbf{do}
    \For{
        \SetKwBlock{If}{}{}\textbf{if} $data\_point\_score > threshold\_score$ \textbf{then}
            \If{
                \lIf{$rand(0,1) \geq swap\_probability$}{\\~~~~swap data point to alternate partition}
                
            }
    }
    \hfill\\
    // Decay threshold\_proportion and swap\_probability for next epoch\\
    $threshold\_proportion$, $swap\_probability$ = $\frac{0.1}{1.1^{epoch}}$\\
    $epoch$ += 1
}
\hfill\\
// Return stratified splits\\
\textbf{return} $X\_train$, $X\_test$, $y\_train$, $y\_test$
\end{algorithm}
\end{tcolorbox}

Finally, a proportion of the data points with the highest scores are swapped from their current partition to the alternate partition (lines 32 to 36). At each epoch, only a proportion of the highest-scored instances get swapped. This proportion is based on the $threshold\_proportion$ and $swap\_probability$, which decays with every epoch (lines 38 to 40). In our experiments, we found that constraining the number of instances that gets swapped is import to mitigate ‘overshooting’. 

The values for $epochs$, $swap\_probability$, $threshold\_proportion$, and $decay$ are key parameters for the provided python implementation of the algorithm. During development, we found that the optimal values depended on the characteristics of each dataset, but very good results were obtained using certain default values. The default values were determined by testing different combinations for all 18 datasets on the XML repository~\cite{ref_url_xmlrepo}. The default value is 10\% for the starting $swap\_probability$ and $threshold\_proportion$. Higher values result in more aggressive stratification, but can result in ‘overshoot’. $swap\_probability$ and $threshold\_proportion$ are $decayed$ at a default rate of 10\% per epoch, and the default number of $epochs$ is 50. We found that these values worked well, and either matched or came very close to the partitions that were generated using tailored values. However, users of the algorithm can easily trial different values by passing in the desired parameter values when calling the function.

One notable behaviour of the algorithm is that it increases the test size as needed to achieve stratified partitions with few missing labels. It automatically finds this test size by balancing two competing priorities: 1) generating partitions where the proportion of label instances in the test set is equal to the $target\_test\_size$, and 2) reducing the number of missing labels from either set. Initially, the algorithm swaps more instances into the test set since the decrease in score achieved from reducing the number of missing labels is greater than the increase in score caused by deviating from the $target\_test\_size$. At some point, the respective changes to the score reaches equilibrium; this is the test size that the algorithm settles on. We believe this to be a desirable trait, since we consider generating stratified partitions with few missing labels to be more important than maintaining an arbitrary test size. However, we acknowledge that this may not be suitable for all applications.

\section{Partitioning Results}

In this section, we compare the test-train splits generated by random, iterative~\cite{ref_article_iterative}, and stratified sampling against the splits provided on the XML repository~\cite{ref_url_xmlrepo}. We generated splits for all 18 datasets and calculated two statistics for each test set: 1) KL-Divergence, and 2) Percentage of labels missing. The results are presented in Table~\ref{tab_fullresults}. The lowest values for each dataset are bolded.

Following the work of Aguilar et al., we measured KL-Divergence of each test set against their respective dataset~\cite{ref_article_lince}. Also known as relative entropy, this metric measures the difference between two probability distributions. While not typically used for this purpose, this metric succinctly conveys the extent to which the label ratios are maintained after partitioning. A high number indicates that the label distributions are highly divergent, while values close to zero indicate they are very similar.

\begin{landscape}
\def\arraystretch{1.37}%
\begin{table}[htbp]
  \centering
  \caption{Label distribution statistics of XML datasets - Statistics are of the test set.}
    \makebox[\textwidth][c]{
    \begin{tabular}{p{11em} |
    >{\centering\arraybackslash}p{4.5em}
    >{\centering\arraybackslash}p{4.5em} |
    >{\centering\arraybackslash}p{5em}
    >{\centering\arraybackslash}p{5em} |
    >{\centering\arraybackslash}p{4.5em}
    >{\centering\arraybackslash}p{4.5em}
    >{\centering\arraybackslash}p{4em} |
    >{\centering\arraybackslash}p{4.5em}
    >{\centering\arraybackslash}p{4.5em}
    >{\centering\arraybackslash}p{4em}}
    \toprule
          & \multicolumn{2}{c|}{\textbf{Provided~\cite{ref_url_xmlrepo}}} & \multicolumn{2}{c|}{\textbf{Random sampling}} & \multicolumn{3}{c|}{\textbf{Iterative algorithm~\cite{ref_article_iterative}}} & \multicolumn{3}{c}{\textbf{Stratified sampling}} \\
    \midrule
          & KL-Divergence & \% labels missing & KL-Divergence & \% labels missing & KL-Divergence & \% labels missing & time (mins) & KL-Divergence & \% labels missing & time (mins) \\
    \midrule
    \rowcolor{gray!10} Mediamill & \textbf{$<$0.001} & \textbf{0.0\%} & 0.001 & \textbf{0.0\%} & \textbf{$<$0.001} & \textbf{0.0\%} & 0.8 & \textbf{$<$0.001} & \textbf{0.0\%} & 0.1 \\
    Bibtex & 0.009 & \textbf{0.0\%} & 0.01  & \textbf{0.0\%} & \textbf{$<$0.001} & \textbf{0.0\%} & 0.03 & \textbf{$<$0.001} & \textbf{0.0\%} & 0.02 \\
    \rowcolor{gray!10} Delicious & 0.008 & \textbf{0.0\%} & 0.006 & \textbf{0.0\%} & 0.005 & \textbf{0.0\%} & 2.4 & \textbf{0.001} & \textbf{0.0\%} & 0.1 \\
    RCV1-2K & 0.023 & 2.1\% & 0.003 & 0.6\% & 0.003 & 0.7\% & 110.7 & \textbf{$<$0.001} & \textbf{0.0\%} & 1.4 \\
    \rowcolor{gray!10} EURLex-4K & 0.602 & 32.4\% & 0.501 & 28.8\% & 0.444 & 29.2\% & 2.5 & \textbf{0.103} & \textbf{12.3\%} & 0.03 \\
    EURLex-4.3K & 0.474 & 40.0\% & 0.4 & 36.5\% & 0.29  & 33.5\% & 5.2 & \textbf{0.066} & \textbf{13.4}\% & 0.1 \\
    \rowcolor{gray!10} AmazonCat-13K & 0.004 & 0.4\% & 0.02  & 10.0\% & 0.007 & 6.0\% & 488.6 & \textbf{0.001} & \textbf{0.2\%} & 3.6 \\
    AmazonCat-14K & 0.001 & \textbf{0.0\%} & 0.006 & 4.7\% & - & - & $>$24H & \textbf{$<$0.001} & 0.1\% & 11.0 \\
    \rowcolor{gray!10} Wiki10-31K & 1.106 & 28.7\% & 0.681 & 18.0\% & 0.648 & 17.3\% & 117.1 & \textbf{0.264} & \textbf{7.2\%} & 0.2 \\
    Delicious-200K & 0.067 & 11.1\% & 0.065 & 10.4\% & - & - & $>$24H & \textbf{0.023} & \textbf{3.6\%} & 16.1 \\
    \rowcolor{gray!10} WikiLSHTC-325K & 0.157 & 12.8\% & 0.563 & 23.1\% & - & - & $>$24H & \textbf{0.103} & \textbf{9.6\%} & 9.2 \\
    WikiSeeAlsoTitles-350K & 0.772 & \textbf{5.5\%} & 2.865 & 31.6\% & - & - & $>$24H & \textbf{0.582} & 8.5\% & 2.4 \\
    \rowcolor{gray!10} WikiTitles-500K & 0.09  & 2.0\% & 0.43  & 12.7\% & - & - & $>$24H & \textbf{0.026} & \textbf{0.9\%} & 12.4 \\
    Wikipedia-500K & 0.036 & \textbf{0.0\%} & 0.399 & 11.5\% & - & - & $>$24H & \textbf{0.018} & 0.1\% & 15.6 \\
    \rowcolor{gray!10} AmazonTitles-670K & 5.757 & 48.6\% & 1.98  & 18.5\% & - & - & $>$24H & \textbf{0.198} & \textbf{1.9\%} & 4.7 \\
    Amazon-670K & 5.714 & 48.2\% & 1.96  & 18.1\% & - & - & $>$24H & \textbf{0.182} & \textbf{1.5\%} & 4.8 \\
    \rowcolor{gray!10} AmazonTitles-3M & 0.033 & \textbf{0.1\%} & 0.215 & 7.7\% & - & - & $>$24H & \textbf{0.031} & 0.8\% & 85.4 \\
    Amazon-3M & \textbf{0.03}  & \textbf{0.0\%} & 0.215 & 7.6\% & - & - & $>$24H & 0.031 & 0.7\% & 79.1 \\
    \bottomrule
    \end{tabular}}%
  \label{tab_fullresults}%
\end{table}%
\end{landscape}

\vspace*{-\baselineskip}
\begin{figure}
\includegraphics[width=1.05\textwidth]{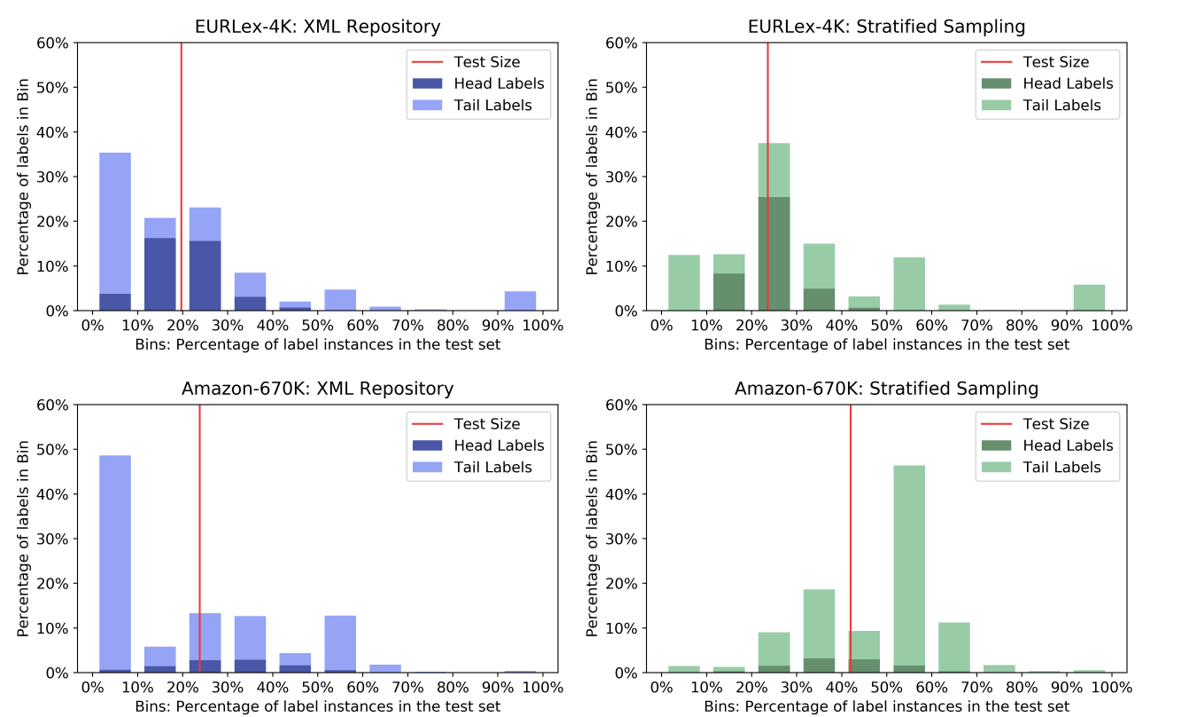}
\caption{Comparison of label distributions between provided and stratified splits.} \label{fig_split_comparison}
\end{figure}
\vspace*{-\baselineskip}

Firstly, we observe that random sampling produced highly divergent partitions for many of the datasets. This is most apparent for the EURLex, Wiki, and very large Amazon datasets, where the KL-Divergences of their test sets are comparatively high. They are also missing a large proportion of labels. This highlights the risk of relying on random sampling for partitioning XML data - it can result in splits that are highly unrepresentative of the dataset. Based on the results, this risk appears to be greatest for very large datasets.

Interestingly, we see that the provided splits are sometimes more-, and sometimes less-, stratified than what was generated from random sampling. For RCV1-2K, EURLex-4(.3)K, Wiki10-31K, and Amazon(Titles)-670K, the provided partitions are materially more divergent. Based on the results, it appears that non-random partitioning methods were used to create the splits. It's not clear to us what they were, or the motivation behind them, since they produced splits with higher KL-Divergence and more missing labels. On the other hand, the partitions for AmazonCat, very large Wiki, and Amazon(Titles)-3M are materially more representative of their dataset than what was produced through random sampling. In these cases, KL-Divergences are much lower, and significantly fewer labels are missing. The authors of the XML repository~\cite{ref_url_xmlrepo} do briefly mention that they used some method to ensure the test sets contain as many labels as possible, but what that method is remains unclear.

In any case, the stratified sampling algorithm produced the best results overall. In all cases but one, it generated splits with the lowest KL-Divergence. It also produced splits with the fewest missing labels from the test set in a majority of cases. However, for the very large datasets, a number of the provided splits contain fewer missing labels. This suggests that the authors' partitioning method prioritised minimizing the amount of missing labels. Meanwhile, the stratified sampling algorithm prioritizes generating representative splits.

Finally, we note that the stratified sampling algorithm outperforms the iterative algorithm~\cite{ref_article_iterative}. It produced better splits overall, and did so more efficiently. For example, for AmazonCat-13K, stratified sampling was completed in 3.6 minutes, while the iterative algorithm took over 8 hours to generate a result. In fact, it was unable to generate a result within 24 hours for the largest datasets.

The label distributions of two datasets are plotted in Figure~\ref{fig_split_comparison}\footnote{Remaining charts available at: \url{https://github.com/maxitron93/stratified_sampling_for_XML/tree/master/label_distribution_charts}}. As in Section 3, each chart shows the proportion of labels that fall in each of the 10 bins, and the red line represents the test size. The charts show that the stratified sampling algorithm produced splits with a greater concentration of labels around the vertical red line, and fewer labels in the first and last bins. This provides visual confirmation of their lower KL-Divergence scores.

\section{Bias in Estimated Performance}

Here, we compare the performance estimates of three XML models trained on several provided and stratified splits. We selected Parabel~\cite{ref_article_parabel}, FastXML~\cite{ref_article_fastxml} and AnnexML~\cite{ref_article_annexml} for testing since they are commonly used as benchmark models. Following prevailing literature, we used Precision@1 (P@1) to represent performance. Training and evaluation was carried out on datasets with a large difference in the amount of missing labels between the provided and stratified splits. We also included two datasets with very little difference between splits. The results are presented in Table~\ref{tab_performanceestimates}.

As we can see, there are material deltas in estimated performance for datasets where there is a large difference between the provided and stratified data splits (EURLex-4(.3)K, Wiki10-31K, Delicious-200K and Amazon-670K). 
For these datasets, it appears that models trained and evaluated on the provided split achieves higher P@1 than if the stratified split was used. On the other hand, there is significantly less delta for RCV1-2K and AmazonCat-13K - datasets where there is little difference between the provided and stratified partitions.

These results highlight how training and testing models using unrepresentative splits can lead to biased performance estimates. It makes sense that excluding tail labels during evaluation results in higher P@1, since they are the hardest-to-train for labels. Conversely, including them would drag down overall performance. This is most apparent for Amazon-670K, where P@1s of the models trained with stratified splits are as much has 22\% lower. This coincides with the very large proportion of labels missing from the provided test set: 48.2\%.

\def\arraystretch{1.3}%
\begin{table}[t]
  \centering
  \caption{Difference in P@1 between models trained and tested using different partitions}
  \begin{adjustbox}{width=\textwidth}
    \begin{tabular}{p{7.4em} |
    >{\centering\arraybackslash}p{4.1em}
    >{\centering\arraybackslash}p{4.1em}
    >{\centering\arraybackslash}p{3.3em} |
    >{\centering\arraybackslash}p{4.1em}
    >{\centering\arraybackslash}p{4.1em}
    >{\centering\arraybackslash}p{3.3em} |
    >{\centering\arraybackslash}p{4.1em}
    >{\centering\arraybackslash}p{4.1em}
    >{\centering\arraybackslash}p{3.3em}}
    \toprule
          & \multicolumn{3}{c|}{\textbf{Parabel}} & \multicolumn{3}{c|}{\textbf{FastXML}} & \multicolumn{3}{c}{\textbf{AnnexML}} \\
          & Provided & Stratified & \%$\Delta$ & Provided & Stratified & \%$\Delta$ & Provided & Stratified & \%$\Delta$ \\
    \midrule
    \rowcolor{gray!10} EURLex-4K & 0.819 & 0.791 & \textbf{-3.5\%} & 0.716 & 0.682 & \textbf{-4.8\%} & 0.799 & 0.775 & \textbf{-3.0\%} \\
    EURLex-4.3K & 0.907 & 0.881 & \textbf{-2.9\%} & 0.868 & 0.835 & \textbf{-3.8\%} & 0.903 & 0.870 & \textbf{-3.7\%} \\
    \rowcolor{gray!10} Wiki10-31K & 0.842 & 0.830 & \textbf{-1.5\%} & 0.828 & 0.815 & \textbf{-1.5\%} & 0.865 & 0.857 & \textbf{-1.0\%} \\
    Delicious-200K & 0.466 & 0.425 & \textbf{-8.8\%} & 0.432 & 0.390 & \textbf{-9.7\%} & 0.465 & 0.429 & \textbf{-7.7\%} \\
    \rowcolor{gray!10} Amazon-670K & 0.447 & 0.373 & \textbf{-17\%} & 0.370 & 0.288 & \textbf{-22\%} & 0.420 & 0.336 & \textbf{-20\%} \\
    \midrule
    RCV1-2K & 0.904 & 0.911 & \textbf{+0.7\%} & 0.912 & 0.916 & \textbf{+0.5\%} & 0.906 & 0.911 & \textbf{+0.6\%} \\
    \rowcolor{gray!10}AmazonCat-13K & 0.934 & 0.933 & \textbf{-0.1\%} & 0.931 & 0.932 & \textbf{+0.1\%} & 0.935 & 0.935 & \textbf{0.0\%} \\
    \bottomrule
    \end{tabular}
    \end{adjustbox}
  \label{tab_performanceestimates}
\end{table}%

In addition to generating biased performance estimates, using unrepresentative splits could also results in sub-optimal choices for hyperparameters - it's plausible to think that what's suitable for the specific split might not be the best choice for the entire dataset. Overall, these results lend more support to the already well established notion that using stratified splits during model development helps with ensuring the derivation of robust performance estimates~\cite{ref_article_bias}~\cite{ref_article_iterative}. As such, stratified splits should be used where available.

\section{Conclusion}

In this paper, we examined the label distributions of test-train splits of XML data, generated using different partitioning methods. We also studied their impact on estimated performance. A summary of our findings is as follows: 

\begin{itemize}
  \item Many of the commonly used test-train splits provided by the XML repository are not representative of the entire dataset. Random sampling is not a suitable alternative since it produced partitions with label distributions that are highly divergent from the entire dataset.
  \item Stratified sampling is commonly used in the binary and multi-class settings, but those methods cannot be applied to multi-label data. Existing methods for MLC don't work well for XML-scale datasets: the commonly used iterative algorithm~\cite{ref_article_iterative} is slow, and cannot produce well stratified splits.
  \item Our proposed stratified sampling method is capable of efficiently generating representative test-train splits of XML data that contain fewer missing labels compared to random sampling and most of the provided splits.
  \item Training and testing models with unrepresentative data can lead to optimistic performance estimates since, currently, many of the hardest-to-train-for tail labels aren't included in a number of the test sets.
\end{itemize}

Overall, our findings lend further support to the notion that it's apt to use stratified subsets of data during model development. As XML research becomes increasingly prevalent in this era of big data, we hope that future researchers will find our stratified sampling method useful in their endeavours. 

\bibliographystyle{splncs04}
\bibliography{pakdd2021}

\begin{thebibliography}{10}
\providecommand{\url}[1]{\texttt{#1}}
\providecommand{\urlprefix}{URL }
\providecommand{\doi}[1]{https://doi.org/#1}

\bibitem{ref_article_lince}
Aguilar, G., Kar, S., Solorio, T.: Lince: A centralized benchmark for
  linguistic code-switching evaluation. In: Proceedings of The 12th Language
  Resources and Evaluation Conference. pp. 1803--1813 (2020)

\bibitem{ref_url_xmlrepo}
Bhatia, K., Dahiya, K., Jain, H., Mittal, A., Prabhu, Y., Varma, M.: The
  extreme classification repository: Multi-label datasets and code (2016),
  \url{http://manikvarma.org/downloads/XC/XMLRepository.html}

\bibitem{ref_article_pfastrexml}
Jain, H., Prabhu, Y., Varma, M.: Extreme multi-label loss functions for
  recommendation, tagging, ranking; other missing label applications. In:
  Proceedings of the 22nd ACM SIGKDD International Conference on Knowledge
  Discovery and Data Mining. p. 935–944 (2016)

\bibitem{ref_article_bias}
Kohavi, R.: A study of cross-validation and bootstrap for accuracy estimation
  and model selection. In: Proceedings of the 14th International Joint
  Conference on Artificial Intelligence - Volume 2. p. 1137–1143 (1995)

\bibitem{lu2020multi}
Lu, J., Du, L., Liu, M., Dipnall, J.: Multi-label few/zero-shot learning with
  knowledge aggregated from multiple label graphs. In: Proceedings of the 2020
  Conference on Empirical Methods in Natural Language Processing. pp.
  2935--2943 (2020)

\bibitem{ref_article_scikitlearn}
Pedregosa, F., Varoquaux, G., Gramfort, A., Michel, V., Thirion, B., Grisel,
  O., Blondel, M., Prettenhofer, P., Weiss, R., Dubourg, V., et~al.:
  Scikit-learn: Machine learning in python. the Journal of machine Learning
  research  \textbf{12},  2825--2830 (2011)

\bibitem{ref_article_parabel}
Prabhu, Y., Kag, A., Harsola, S., Agrawal, R., Varma, M.: Parabel: Partitioned
  label trees for extreme classification with application to dynamic search
  advertising. In: Proceedings of the 2018 World Wide Web Conference. p.
  993–1002 (2018)

\bibitem{ref_article_fastxml}
Prabhu, Y., Varma, M.: Fastxml: A fast, accurate and stable tree-classifier for
  extreme multi-label learning. In: Proceedings of the 20th ACM SIGKDD
  International Conference on Knowledge Discovery and Data Mining. p. 263–272
  (2014)

\bibitem{ref_article_iterative}
Sechidis, K., Tsoumakas, G., Vlahavas, I.: On the stratification of multi-label
  data. In: Proceedings of the 2011 European Conference on Machine Learning and
  Knowledge Discovery in Databases - Volume Part III. p. 145–158 (2011)

\bibitem{ref_article_mlc1}
Shaheen, Z., Wohlgenannt, G., Filtz, E.: Large scale legal text classification
  using transformer models. arXiv preprint arXiv:2010.12871  (2020)

\bibitem{ref_article_network}
Szyma{\'n}ski, P., Kajdanowicz, T.: A network perspective on stratification of
  multi-label data. In: First International Workshop on Learning with
  Imbalanced Domains: Theory and Applications. pp. 22--35 (2017)

\bibitem{ref_article_multilearn}
Szymanski, P., Kajdanowicz, T.: Scikit-multilearn: A scikit-based python
  environment for performing multi-label classification. J. Mach. Learn. Res.
  \textbf{20}(1),  209–230 (2019)

\bibitem{ref_article_annexml}
Tagami, Y.: Annexml: Approximate nearest neighbor search for extreme
  multi-label classification. In: Proceedings of the 23rd ACM SIGKDD
  International Conference on Knowledge Discovery and Data Mining. p. 455–464
  (2017)

\bibitem{ref_article_mulan}
Tsoumakas, G., Spyromitros-Xioufis, E., Vilcek, J., Vlahavas, I.: Mulan: A java
  library for multi-label learning. The Journal of Machine Learning Research
  \textbf{12},  2411--2414 (2011)

\bibitem{ref_article_mlc2}
Wagner, P., Strodthoff, N., Bousseljot, R.D., Kreiseler, D., Lunze, F.I.,
  Samek, W., Schaeffter, T.: Ptb-xl, a large publicly available
  electrocardiography dataset. Scientific Data  \textbf{7}(1),  1--15 (2020)

\bibitem{ref_article_attentionxml}
You, R., Zhang, Z., Wang, Z., Dai, S., Mamitsuka, H., Zhu, S.: Attentionxml:
  Label tree-based attention-aware deep model for high-performance extreme
  multi-label text classification. In: Advances in Neural Information
  Processing Systems 32. pp. 5820--5830 (2019)

\end{thebibliography}

\end{document}